\documentclass[journal]{IEEEtran}
\usepackage{graphicx}
\usepackage{amssymb}
\usepackage{array}
\usepackage{caption}
\usepackage{subcaption}
\let\labelindent\relax
\usepackage{enumitem}
\usepackage{color,soul}
\usepackage[numbers,sort&compress]{natbib}

\newcommand{\TangoPlus}{TangoPlus\texttrademark~}
\newcommand{\VeroClear}{VeroClear\texttrademark~}
\newcommand{\Printer}{Stratasys\textregistered~ Connex 500\texttrademark~}
\title{Design and locomotion control of soft robot using friction manipulation and motor-tendon actuation}
\usepackage{setspace}

\begin{document}
\author{Vishesh~Vikas, %
			Eliad~Cohen, Rob~Grassi, Canberk~S\"ozer~%
			and~Barry~Trimmer

\thanks{This work was funded in part by the National Science Foundation grant IOS-1050908 to Barry Trimmer and National Science Foundation Award DBI-1126382. \newline \indent
Vishesh Vikas, Rob Grassi and Barry Trimmer are with the Neuromechanics and Biomimetic Devices Laboratory, Tufts University, Medford MA 02155 (email: vishesh.vikas@tufts.edu, robert.grassi@tufts.edu and barry.trimmer@tufts.edu )
Eliad Cohen is with University of Massachusetts, Lowell (eliad.cohen@gmail.com)
Canberk S\"ozer is with Yildiz Technical University, Istanbul 34330, Turkey (email: canberksozer@gmail.com)
} 
}

\IEEEpeerreviewmaketitle

\maketitle
\begin{abstract}
Robots built from soft materials can alter their shape and size in a particular profile. This shape-changing ability could be extremely helpful for rescue robots and those operating in unknown terrains and environments. In changing shape, soft materials also store and release elastic energy, a feature that can be exploited for effective robot movement. However, design and control of these moving soft robots are non-trivial. The research presents design methodology for a 3D-printed, motor-tendon actuated soft robot capable of locomotion. The modular design of the robot facilitates rapid fabrication, deployment and repair. In addition to shape change, the robot uses friction manipulation mechanisms to effect locomotion. The motor-tendon actuators comprise of nylon tendons embedded inside the soft body structure along a given path with one end fixed on the body and the other attached to a motor. These actuators directly control the deformation of the soft body which influences the robot locomotion behavior. Static stress analysis is used as a tool for designing the shape of the paths of these tendons embedded inside the body. The research also presents a novel model-free learning-based control approach for soft robots which interact with the environment at discrete contact points. This approach involves discretization of factors dominating robot-environment interactions as states, learning of the results as robot transitions between these robot states and evaluation of desired periodic state control sequences optimizing a cost function corresponding to a locomotion task (rotation or translation). The clever discretization allows the framework to exist in robot's task space, hence, facilitating calculation of control sequences without modeling the actuator, body material or details of the friction mechanisms. The flexibility of the framework is experimentally explored by applying it to robots with different friction mechanisms and different shapes of tendon paths.
\end{abstract}
%
\section{Introduction}
{\LARGE I}\textsc{nsights} from studies of animal movements have inspired a variety of deformable robots \cite{hirose_biologically_2004, wright_design_2007, chirikjian_kinematics_1995}, many with intended applications in search and rescue \cite{erkmen_snake_2002}. These snake-like serpentine robots consisting of rigid articulated segments have been researched for maneuvering through complex unpredictable environments. An alternative approach is to build machines from soft materials making them more robust and adaptable. Soft materials are comparatively rare in robots but have been used for manipulation, gripping and locomotion  \cite{laschi_design_2009, cheng_design_2012, brown_universal_2010, seok_peristaltic_2010, shepherd_multigait_2011, umedachi_highly_2013}. The soft, deformable properties of continuum materials \cite{pfeifer_morphological_2005} are potentially useful for overcoming limitations posed by traditional hard, rigid body robotic systems because they allow the structure to change shape by twisting, bending and generally deforming in all dimensions. These properties make them ideal for search and rescue operations in unknown environments. With such a future application in mind, this research explores design and locomotion control of a soft body robot in structured planar environment. Soft robots create their own challenges with design and control. 

\textit{Design Challenges.} The design challenges for soft robots can be classified into fabrication methodology, material selection, soft actuation and 
friction mechanisms. Here, the friction challenges are unique for locomotion requirements. Soft robots are typically fabricated using casting techniques \cite{cianchetti_design_2011,lin_goqbot:_2011,shepherd_multigait_2011}. Casting techniques are very powerful and allow use of materials with wide range of mechanical properties. However, multi-material manufacturing is very difficult when casting into molds and requires either insert-molding or two stage casting systems that make tooling more complicated. The manufacturing technique of controlled layer-by-layer polymer deposition, referred to as additive manufacturing (3D printing), has the ability to provide solutions to soft robot manufacturing by providing flexibility of simultaneously using multiple materials. More importantly, the additive manufacturing methods of Fused Deposition Modeling (FDM) and Stereolithography (SLA) are available as off-the-shelf platforms from different companies. FDM is a method where a solid polymer filament is liquefied by heating and deposited to form a solid layer of the manufactured part. In contrast, SLA employs a liquid pre-polymer that is patterned onto a surface and cured (e.g. using UV exposure) to form a solid layer of the manufacturing part. While allowing fast turnaround and complex multi-material designs, additive manufacturing limits the material choices available to the designer as most printers require proprietary resins to be used. It is possible, with printers like \Printer ~to create materials with properties ranging between rigid and elastomeric by blending the two resins (an out of the box software feature) but the possibilities are still within the realm of the available resins. %
The limitation of material selection with additive manufacturing can be overcome upon finalization of the robot design by manufacturing that design using casting techniques. The soft robot described in this research utilizes multi-material additive manufacturing technique on the \Printer using soft-rigid materials provided for design purposes.

Actuation of soft robots remains a challenge because most electromagnetic systems are made of rigid materials \cite{kim_soft_2013}. Alternative systems such as dielectric elastomeric actuators (DEAs) have been explored \cite{ohalloran_review_2008,carpi_biomedical_2009} but these require high voltages for actuation and, without a rigid frame, produce very low stress \cite{cianchetti_new_2009}. Other flexible actuators including pressurized liquid or air, shape memory alloys (SMAs) coils activated by joule heating have also been explored \cite{daerden_pneumatic_2002,wehner_experimental_2012,%
shepherd_multigait_2011,menciassi_development_2006,lin_goqbot:_2011,%
laschi_design_2009, kim_soft_2013, umedachi_highly_2013}. In such cases, the actuation is either slow or unpredictable. Motor controlled cable-driven systems have been researched for manipulation, hands and as a secondary actuation for locomotion in water \cite{laschi_design_2009, cieslak_elephant_1999, webster_design_2010, walker_continuum_2005, mcmahan_design_2005,controzzi_bio-inspired_2010,calisti_octopus-bioinspired_2011,marques_ecce1:_2010}. A soft robot capable of terrestrial locomotion that uses a motor-tendon system is described in this research. The motors are located in mountings that do not interfere with movements of the soft body itself. The tendons are embedded inside the soft robot body where they act as both actuators and structural components. The path and placement of these tendons are a fundamental design feature affecting how the robot body deforms upon actuation.

A major factor in effective terrestrial locomotion is the interaction between the body and its environment. The forces required to initiate or maintain differential movement between interacting surfaces are often dominated by friction. Animals have evolved a variety of mechanisms to exploit these forces including directionally sensitive friction, chemical or electrostatic adhesives, structures that exploit asperities in different size ranges and deployable or retractable grippers \cite{autumn_frictional_2006,creton_sticky_2007,%
gorb_attachment_2001,mezoff_biomechanical_2004, belanger_combined_2000}. In general, terrestrial locomotion results from spatial and temporal control of frictional forces \cite{radhakrishnan_locomotion:_1998}. The non-wheel options for manipulating friction range from gecko-like friction adhesive \cite{autumn_frictional_2006}, anisotropic friction mechanism \cite{koh_omega-shaped_2013}, microspines \cite{spenko_biologically_2008, asbeck_scaling_2006, parness_gravity-independent_2013}, micro-hooks \cite{menciassi_development_2006} to electromagnets \cite{kotay_inchworm_2000}, jamming \cite{brown_universal_2010} and whegs \cite{schroer_comparing_2004}. Simplistically, locomotion is a result of minimizing frictional forces at one end of the robot while maximizing it at the other which requires relative difference in friction - directional or grip. 
Although friction control can be accomplished using additional actuators, in this research, the control of the friction mechanisms relies on the shape changes that occur during soft robot movements.

\textit{Control Challenges.} One of the most difficult aspects confronting the design and successful deployment of soft robots is the complexity of their movements in response to applied forces. In addition to static load deformations, soft structures may exhibit time-dependent variations in mechanical properties and undergo instability transformations (e.g., buckling), not as modes of failure but as part of their normal operation. These considerations, together with the difficulty of modeling friction in real world settings, make simulation and physics-based control systems very hard to implement in real time. %
We wish to produce locomotion in these robots by executing periodic control sequences that draw analogy to locomotion gaits. Consequently, we attempt to define a framework that exists in task space (independent of soft material properties, type of actuator) and allows calculation of periodic control sequences. The presented model-free control framework discretizes the factors dominating the frictional interaction with the environment, learns from these interactions and calculates optimal periodic control sequences for locomotion (translation and rotation). This has been implemented to control locomotion in soft robots with different friction mechanisms, on different locomotion surfaces and different shapes of tendon paths.

\textit{Contributions.} The research presents \textbf{design and control methodology} for a 3D printed motor-tendon actuated soft body robot. 
\begin{enumerate}
\item Design methodology (Section \ref{Sec:RobotDesign}) discusses challenges of fabrication, actuation and friction manipulation.
\item Control methodology presents a model-free control framework (Section \ref{Sec:ModelFreeControl}) applicable to soft robots with discrete contact points. The framework allows calculation of periodic optimal locomotion sequences. The flexibility of this control framework is experimentally explored (Section \ref{Sec:Experiments}) by applying it to robots with different friction systems, locomotion surfaces and shapes of tendon paths (discussed later). 
\end{enumerate}
\section{Robot Design} \label{Sec:RobotDesign}
The design space for soft bodied robots is very big. This robot design builds on an existing work of soft-bodied robots inspired by \textit{Manduca sexta} caterpillar locomotion \cite{umedachi_highly_2013}. The previous versions relied on shape memory alloys (SMA) actuation which resulted in slow-moving but hard to control soft robots. The design objectives can be encapsulated as follows
\begin{enumerate}[leftmargin=*]
\item {Deformable body.} 
\begin{enumerate}
\item \textit{Anisotropic bending.} The requirement of the robot body is axial deformation and not lateral deformation (like a caterpillar) and elastic recovery of the body after deformation (energy storage).
\item \textit{Discrete contact points.} Unlike undulatory motion, the robot has  only two contact points with the environment upon locomotion. Each of these contact point interacts with the environment using a friction manipulation mechanism (see below). 
\item \textit{Modular design.} A modular design should facilitate easy repair of actuator failure.
\end{enumerate}
\item {Controlled friction manipulation.} 
\begin{enumerate}
\item \textit{Gripper-like friction mechanism.} The design of a mechanism with grip-like properties that can be engaged/disengaged during the locomotion sequence. 
\item \textit{Directional friction mechanism.} The design of mechanisms which will require variation in friction force with direction of movement.
\end{enumerate}
\item {Actuation.} 
\begin{enumerate}
\item \textit{Precision and fast actuation with low voltage.} The original design was SMA-actuated. They have slow actuation cycle and have been found to give unreliable performance in spite of functioning at low voltage. The challenges arise as their performance depends on heat transfer (temperature-dependent).
\item \textit{Embedding tendons.} Embedding tendons inside the soft body needs to compensate for differences in hardness during interaction between the hard tendons and the soft body upon actuation and will result in tendons damaging the robot body. \item \textit{Tendon path shapes.} The embedded tendon can be
designed to follow any shape. Different shapes of tendon paths induce different body deformations, thus, different results of actuation and locomotion. 
\end{enumerate}
\end{enumerate}
\subsection{Body Design}
\label{Subsec:BodyDesign}
\begin{figure}
\includegraphics[width=\columnwidth]{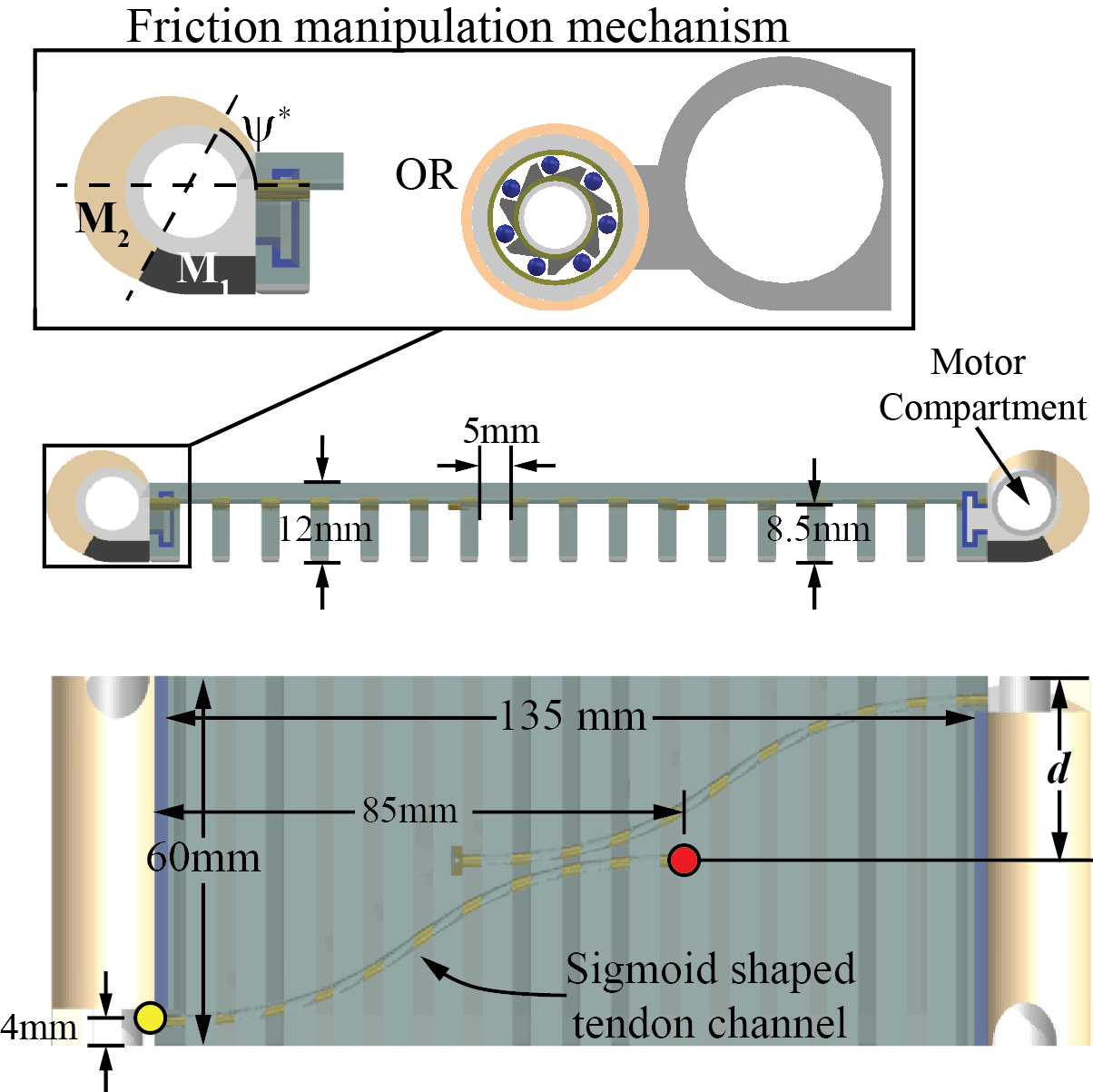}
\caption{Detailed description of the soft robot design. The $135mm\ \times\ 60mm$ rectangle (top view) shaped soft body is attached to friction manipulation mechanisms at each end of the robot. The $8.5mm$ deep grooves impart anisotropy to the otherwise isotropic material and facilitate more deformation about the length of the robot as compared to the width. The friction manipulation mechanism includes a compartment to hold the motor of the motor-tendon actuator. The sigmoid shaped tendon paths start at one edge at $4mm$ distance from the edge (yellow circle). The tendon paths terminate at $85mm$ along the length and $d\ mm$ from the opposite edge (red circle). The blue contour between the robot body and friction mechanism indicates its modular assembly with the soft body.}
\label{Fig:PhotoWithDimensions}
\end{figure}

The robots described here are printed on a \Printer printer utilizing PolyJet\texttrademark~ additive manufacturing technology using \TangoPlus as the soft material (Shore Hardness A 26-28) and \VeroClear (Shore Hardness D 83-86) as the hard material \cite{_http://www.stratasys.com/_????}. The experimental robot has a soft rectangular body with horizontal ribs that impart structural anisotropy to facilitate bending and reduce the substrate contact area (Fig. \ref{Fig:PhotoWithDimensions}). \TangoPlus has a maximum extensibility of approximately $200\%$. Hence, these ribs reduce the local strain experienced during extreme bending and prevent elongation failure. 

\begin{figure}
\centering
\includegraphics[width=\columnwidth]{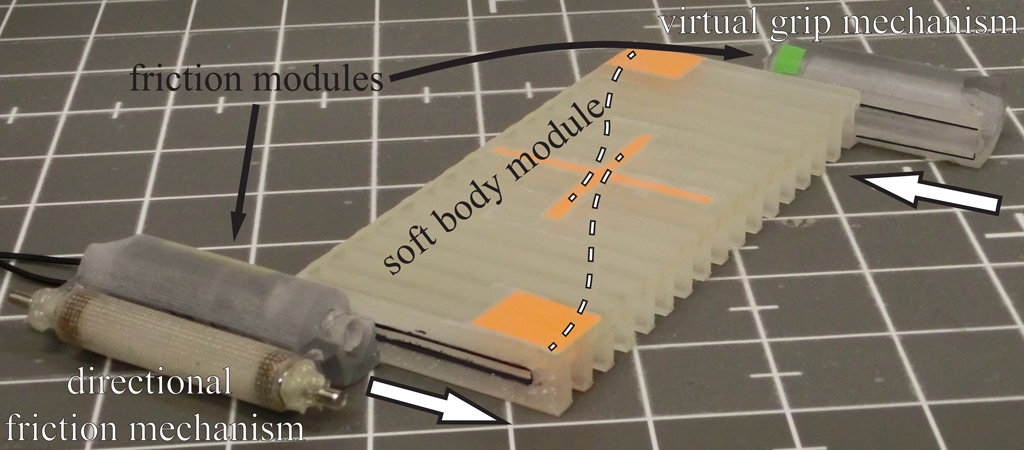}
\caption{The soft robot printed as three modules of two friction mechanisms (for each end) and a soft body. The hollow arrows indicate the direction of movement for the friction modules along the sleeve to attach with the body module. The dotted hollow lines indicate the tendon paths. The motors are inserted inside the friction manipulation mechanism with a pulley attached to the shaft.}
\label{Fig:Module}
\end{figure}

The presented soft robot uses two overlapping, motor-tendon actuators that allow controlled deformation of the soft body. Each motor-tendon actuator consists of a brushless Maxon motor (RE10 256102 with gear head GP10 218416) that actively shortens a Nylon tendon (fishing line) by winding it around a pulley. The distal end of the tendon is attached to the soft robot body (Fig. \ref{Fig:PhotoWithDimensions}). The resting length of the tendon is restored passively by the release of the stored elastic energy as the body relaxes. A design challenge results from the modulus and strength mismatch resulting from interaction between the nylon tendons of the actuator and the soft material used for the robot body. This is solved by internally coating the tendon paths with segments of hard material and allows for both - efficient energy transfer to induce body deformation as well as protection of the soft body from being damaged by the tendons. The robots are designed in a modular fashion - the friction mechanisms and soft body are separately printed. The resulting robot is an assembly of the friction mechanism and soft body snapped together as visible in Fig. \ref{Fig:Module}. This modular design facilitates quicker assembly (human cost) and easier repair. Multi-material printing is instrumental in quick manufacturing of such complex design solution.

\subsection{Friction manipulation design} \label{Subsec:FrictionManipulationDesign}
The friction manipulation is performed using two strategies by utilizing 1) the relative difference in frictional interaction of two different materials with the environment, or 2) the direction of motion to vary the amount of frictional drag between the robot and the surface.

\begin{figure}
\centering
\begin{subfigure}[b]{0.5\textwidth}
\centering
\includegraphics[width=0.8\columnwidth]{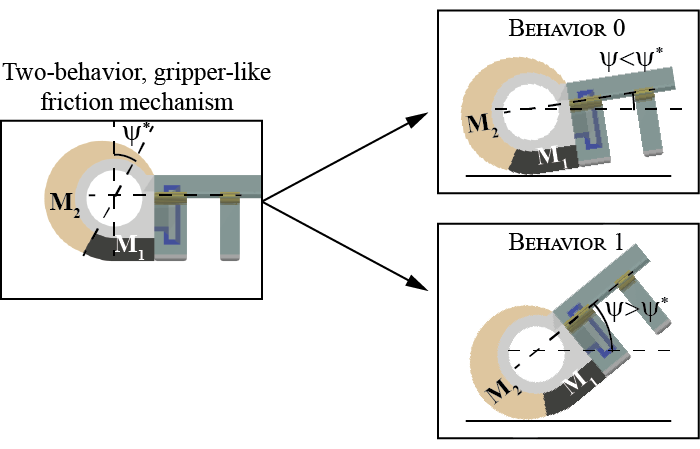}
\caption{Side view of the virtual grip mechanism is made of hard (blue) and soft (red) materials having different coefficients of friction. The frictional force acting on the mechanism changes as the robot shape changes, the mechanism rotates and the material of contact changes about critical contact angle ($\psi^*$).}
\label{Fig:VirtualGrip}
\end{subfigure}
\begin{subfigure}[b]{0.5\textwidth}
\centering
	\includegraphics[width=0.75\columnwidth]{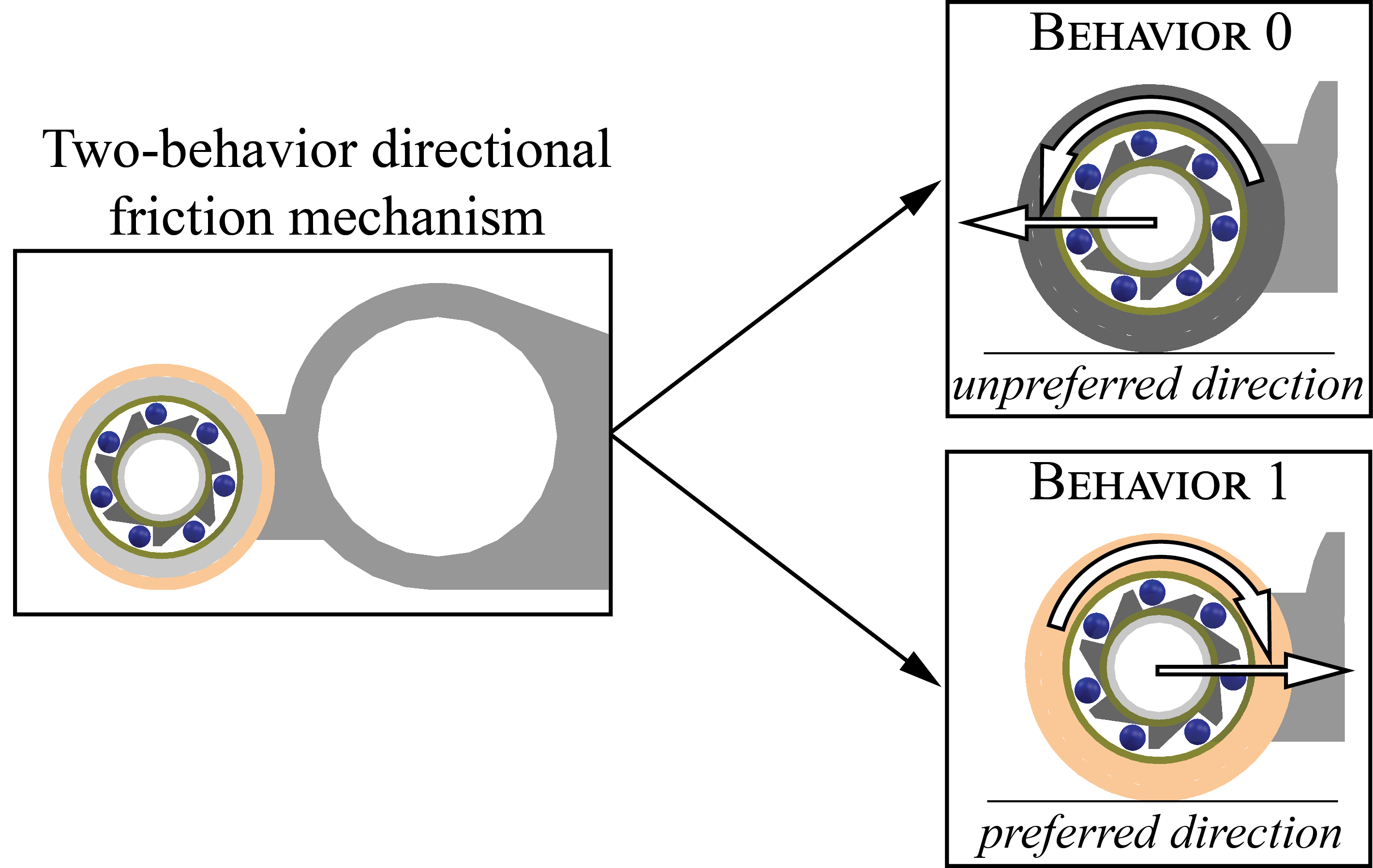}
	\caption{Side view of the unidirectional clutches (grey) are free to rotate (less friction – blue roller pad) in a preferred direction but lock in the opposite direction (more friction – red roller pad). The direction of rotation of the clutch (dotted arrows) determines the direction of translation ($v$) of the mechanism.}
	\label{Fig:DirectionalFriction}
\end{subfigure}
\caption{Friction manipulation mechanisms}
\end{figure}

\textit{Virtual grip mechanism.} This mechanism utilizes relative difference in frictional interaction of two different materials with the environment. It is similar to a variable friction mechanism first described for soft robots powered by shape-memory alloy coils \cite{umedachi_highly_2013} and consists of stiff capsules at each end of the robot that also serve as motor housings (Fig.~\ref{Fig:PhotoWithDimensions}). Each capsule is made from two different materials, one relatively soft ($M_1$, Shore Hardness A 26-28) and the other hard ($M_2$, Shore Hardness D 83-86). The soft material has more friction ($\mu_{static,soft}=0.68$) for an applied load than the harder material ($\mu_{static,hard} = 0.28$). The material in contact with the surface is dependent on the angle of contact ($\psi$). The angle of contact $\psi$ is defined as the angle between the tangent of the soft robot body at the friction mechanism and the surface as shown in Fig. \ref{Fig:VirtualGrip}. Similarly, the critical angle of contact ($\psi^*$) is the angle when the material of contact changes from $M_1$ to $M_2$. As the robot deforms, the point of contact between the robot and the surface changes from one material to another, thus, varying the frictional force acting at that end of the robot. This transition of relative change in friction is similar to that of a gripper, such as a spring clip or clothes-peg that has a very high friction when closed and zero friction when it is opened. Hence, this mechanism is termed as a virtual grip mechanism.

\textit{Directional friction using unidirectional clutches.} Unidirectional clutches, also known as freewheel, rotate free (transmit torque) in one direction (positive velocity – $v>0$) but lock (don't transmit torque) when rotated in the other direction (negative velocity – $v<0$). Therefore, movement in the negative direction is opposed by frictional forces between the substrate and roller (which can be increased by using soft or rough materials on the roller surface) whereas movement in the positive direction is opposed by the very low friction forces inside the bearing (Fig. \ref{Fig:DirectionalFriction}). 
\subsection{Actuator Design} \label{Subsection:ActuatorDesign}

\begin{figure}
\includegraphics[width=\columnwidth]{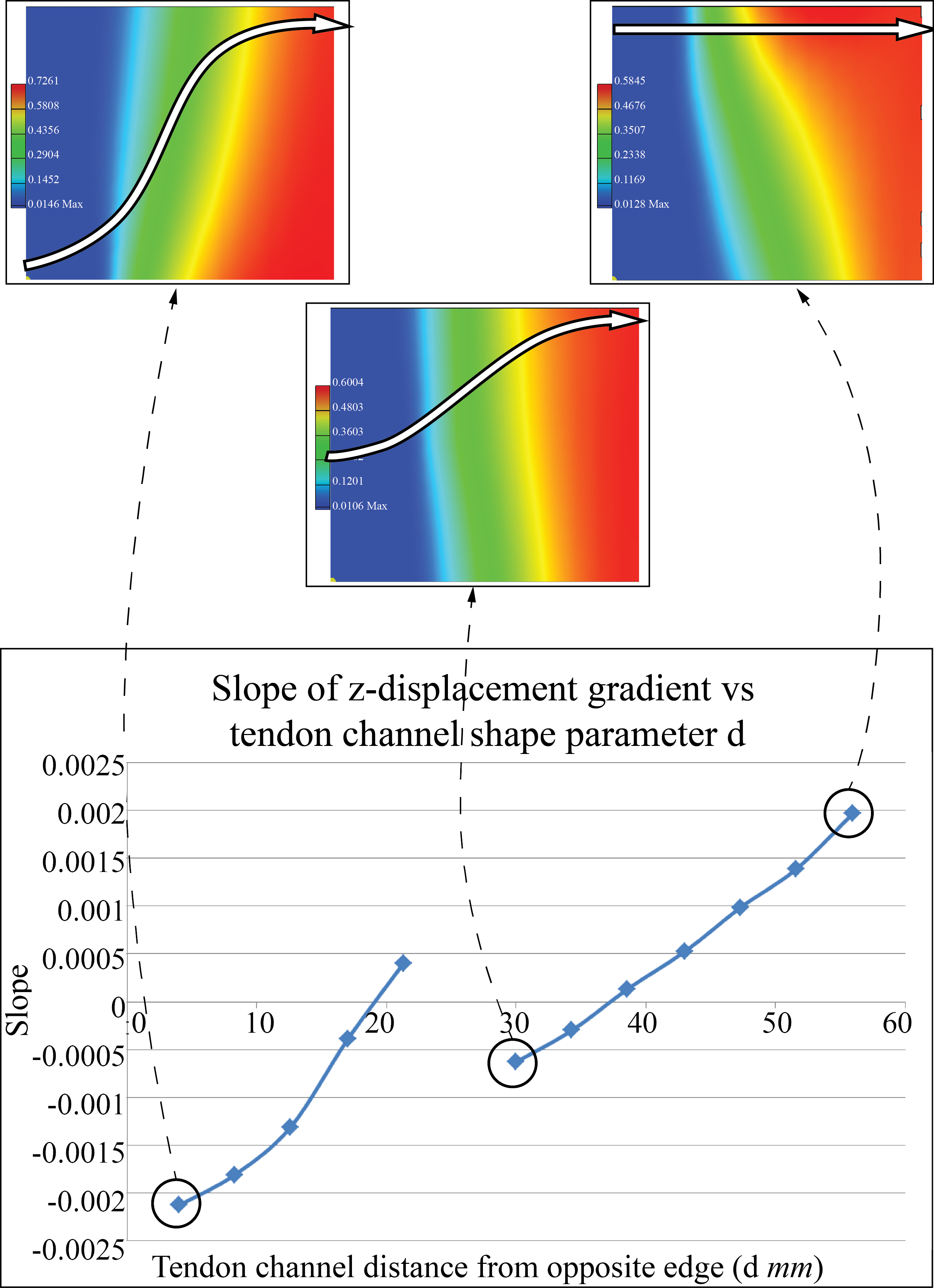}
\caption{Plot of the simulated z-displacement gradient linear slope for different simulated tendon-path designs defined by the distance from the opposite edge $d$. The three circles correspond to the gradient slope for the three specific configurations corresponding to I-shape ($d=56mm$), Midline S-shape ($d=30mm$) and Simple S-shape ($d=4mm$). The tendon path shapes are indicated by white curves where the arrows the end point of the tendon path along the robot edge. These are used for designing tendon path shapes.}
\label{Fig:SlopesAndVariation}
\end{figure}

\begin{figure}
\begin{subfigure}[b]{0.23\textwidth}
\centering
\includegraphics[width=\columnwidth]{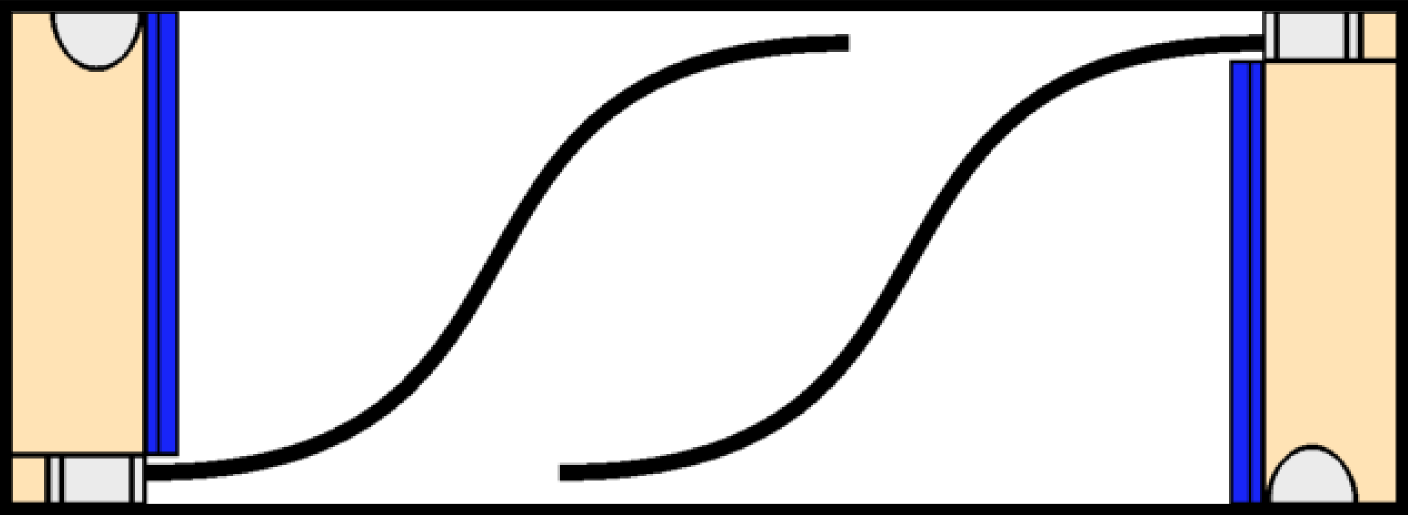}
\caption{Simple S-shape, $d=4mm$}
\label{Fig:SimpleSshapeConfig}
\end{subfigure}
\begin{subfigure}[b]{0.23\textwidth}
\centering
\includegraphics[width=\columnwidth]{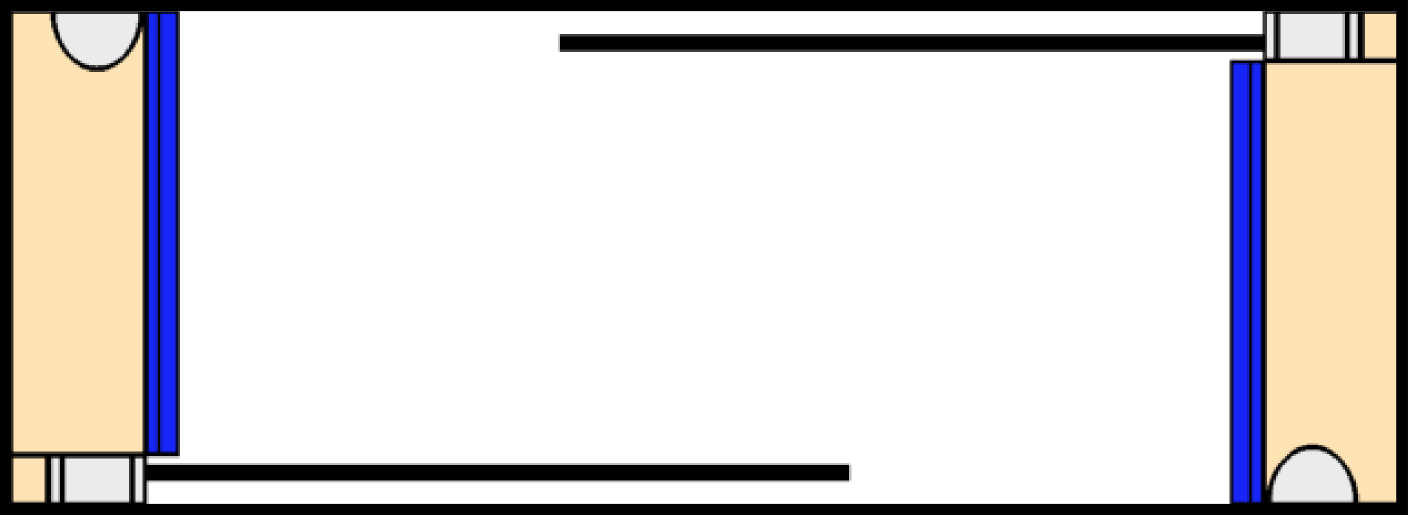}
\caption{I-shape, $d=56mm$}
\label{Fig:IshapeConfig}
\end{subfigure}
\centering
\begin{subfigure}[b]{0.23\textwidth}
\centering
\includegraphics[width=\columnwidth]{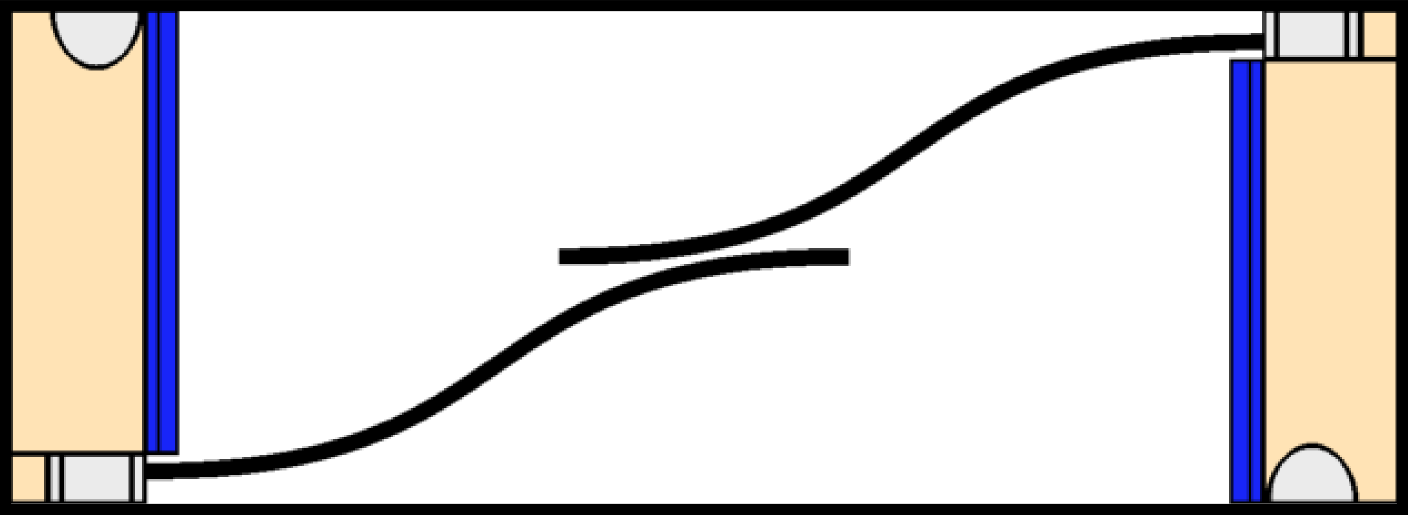}
\caption{Midline S-shape, $d=30mm$}
\label{Fig:MidlineSshapeConfig}
\end{subfigure}
\caption{The three tendon channel shapes experimentally explored with the robot. The two overlapping tendon channels deform the soft body independently control the each friction mechanisms. The robots for $d=56mm,\ 30mm,\ 4mm$ are referred to as the I-shape, Midline S-shape and Simple S-shape soft robots respectively.}
\label{Fig:ManufacturedRobotConfiguration}
\end{figure}

As discussed in Sec. \ref{Subsec:BodyDesign}, the robots are motor-tendon actuated such that the tendons are embedded inside the soft body. The tendon-soft body modulus disparity is overcome by a design conceptually similar to that of bowden cables which have protective housing covering the inside wires. The motor-tendon actuation facilitates fast large deformations using low voltage and precise length-control. The tendon can be designed to pull directly on any part of the robot body and the path of the tendon channel determines how the robot deforms and interacts with the environment, and thereby influences locomotion. In particular, the path shape dependent body deformation and non-symmetric placement (weight distribution) of the motors produce a normal force gradient (assumed proportional to friction) along the width of the robot. The static stress analysis solver of Autodesk Inventor\texttrademark ~2013 is used as a design tool to approximate the normal force gradient profile by assuming it to be proportional to the displacement gradient in the direction normal to the ground (Z-direction). In the analysis, the tendons are modeled as solid low modulus flexible rods constrained to slide along the tendon path inside the body. The length of the tubes does not change but the coiling  motion is simulated by permitting the tendon tubes to extend beyond the robot edge. The attachment point of the tendon (parameter $d$) is varied from $4mm$ (`Simple S-shape') through $30mm$ (`Midline S-shape') to $56mm$ (`I-shape') from the edge of the body to produce a total of 12 simulations for different tendon-channel path configurations. %
The simulated Z-direction displacement profile along the width of the surface contact friction mechanism (as illustrated in Fig. \ref{Fig:SlopesAndVariation}) suggest that friction asymmetry can be programmed into the robot design. The linear slope of the displacement changes with the tendon configuration, varying as a function of the tendon attachment position distance $d$ from the edge. The two extreme configurations of the `Simple S' and `I' shape tendons have a similar effect on deformation but in opposite directions. This means that for the `Simple S' configuration, the z-displacement will be primarily on the pulley side of the motor (marked as arrow in Fig. \ref{Fig:SlopesAndVariation}) whereas in the case of `I' configuration, the z-displacement will peak at the rear of the motor. The design of tendon path shapes is done by using the predicted friction force gradient from the simulation analysis coupled with the assumption that these forces are primary cause of rotation. This simulation design tool suggests that the rotation can be minimized by attaching tendons approximately $37mm$ from the opposite edge (a zero force gradient). Consequently, three variations of tendon shapes are manufactured with attachment points $4mm,\ 30mm$ and $56mm$ from the opposite edge (Figure \ref{Fig:ManufacturedRobotConfiguration}). It is expected that the Midline S-shape ($30mm$) configuration rotates least, I-shape generates counter-clockwise rotation, and robots with the Simple S-shape tendon path have exaggerated clockwise rotation. 
%
\section{Model-Free Control} \label{Sec:ModelFreeControl}
The control of soft/flexible robots may be performed using model-based or model-free control approaches \cite{rigatos_model-based_2009}. The former approach will comprise of modeling of the physical robot, the actuators and the robot-environment interaction. Physical soft robot modeling has been researched using lumped modeling \cite{menciassi_development_2006, saunders_modeling_2011}, continuum modeling \cite{webster_design_2010} and finite element methods \cite{duriez_control_2013}. These approaches tend to ignore the soft continuum properties, are restricted to specific shapes or computationally expensive. The modeling of actuators varies with the type of actuator e.g. SMA modeling is complex, however, motor-tendon is relatively easier and precise. The modeling of the robot interaction with the environment, e.g. friction, use of different friction mechanisms, is difficult \cite{dermitzakis_modeling_2012}. It becomes even more difficult for variable, semi-structured or unstructured environments. The model-free approaches in modeling flexible, soft link robots rely on recurrent neural networks \cite{rigatos_model-based_2009} and reservoir computing \cite{nakajima_exploiting_2014} for learning dynamics of the robot.

For moving soft robots, it is desired to explore an alternative control approach which is generic, adaptable and allows calculation of periodic control sequences. The generic nature of the approach, that allows it to be applicable for different types of actuations, friction manipulation mechanisms and soft materials, can be realized if it exists in task space rather than actuator space \cite{webster_design_2010}. The adaptability of the approach to variable, unknown and unstructured environments can be accomplished through a learning approach. A novel learning-based model-free control is presented that identifies key factors that dominate robot-environment interaction and discretizes these factors into a finite number of states that together capture the overall behavior of the robot. %
This approach learns the results of individual state transitions that may vary as the robot-environment interactions change. Thereafter allowing calculation of optimized periodic control sequences. These periodic control sequences, that may be interpreted as gaits, are expected to be the basic building blocks for robot locomotion. A locomotion task is achieved by implementing the actuator-independent periodic state sequences. The approach can be summarized as follows
\begin{enumerate}
\item \textit{Discretization.} Discretizing the key factors that dominate robot-environment interaction. In process, defining finite robot states and transition reward matrix.
\item \textit{Learning.} For a given state-to-state transition, the rewards are defined as the weighted result of the change in the displacement and orientation of the robot. These rewards are specific to robot design and the locomotion surface.
\item \textit{Optimization.} Calculation of all periodic control sequences. Followed by computation of optimal periodic control sequence corresponding to a locomotion task (translation or rotation) specific cost function.
\end{enumerate}
\subsection{Discretization} \label{Subsec:Discretization}
The described soft robot interacts with the environment at discrete points during locomotion - the friction manipulation mechanisms. These interactions dominate the robot control by influencing the robot-environment interactions and can be discretized as binary behaviors for the virtual grip ($B_{VG}$) or the directional friction mechanism ($B_{DF}$) representing high or low friction
\begin{eqnarray} \label{Eqn:SVG}
B_{VG} &=& \left\{
\begin{array}{cc}
0 &  \psi \leq \psi^* \ (M_1\ in\ contact)\\
1 & \psi > \psi^* \ (M_2\ in\ contact)
\end{array}
\right.%
\\ \label{Eqn:SDF}
B_{DF} &=& \left\{
\begin{array}{cc} 
0 & v>0\ or\ \mathrm{unpreferred\ direction}\\
1 & v<0\ or\ \mathrm{preferred\ direction\phantom{un}}
\end{array}
\right.
\end{eqnarray}
Although two described friction systems are mechanically different, their effects are similar and they both produce behaviors that can be modeled as two discretized binary states indicated in Eq. \ref{Eqn:SVG}, \ref{Eqn:SDF}. Upon discretization, the number of control parameters is same as the number of friction mechanisms (two) i.e. the critical contact angle ($\psi^*$) or the velocity/direction of motion($v$) determine the state of the robot.
%
\subsection{Learning} \label{Subsection:Learning}
The test soft robots have two friction manipulation mechanisms - one at each end and each actuator provides independent control of the frictional mechanism behavior on each end of the robot. The robot is identified as a state $S=(B_1 B_2)$ where $B_1,B_2$ correspond to behaviors of friction mechanisms at each end of the robot. For the present case, the robot can exist in one of the four possible states $(B_1 B_2 )=\{(00),(01),(10),(11)\}$. Additionally, each actuator independently controls a frictional mechanism, hence, allowing the robot to transition from any one of the states to another. As the robot transitions from one state to another, it interacts with the environment which results in some translational and rotational motion. This result of the interaction i.e. change in position $(\Delta x,\Delta y)$ and orientation $(\Delta \theta )$ is weighted and is referred to as the reward. This reward is stored in three state transition reward matrices corresponding to the motions – translation reward matrices ($T_x,T_y$) and rotation reward matrix ($T_\theta$ ). Each state transition reward matrix is of dimension $P^M$ where $P,M$ are the number of discretized behaviors and the number of friction mechanisms respectively. For the present case - $P=2$ (behaviors $0,1$) and $M=2$. Consequently, the $4\times 4$ state transition reward matrix ($T$) stores the weighted result of the interaction of the robot with the environment - the element $T_{(i,j)}$ represents the reward for transition from state  $\mathrm{dec2bin}(i-1)\rightarrow \mathrm{dec2bin}(j-1)$. For the present case, the rewards follow the given codification - displacements of the center of mass in the positive axis, negative axis or none are recorded as $+$ve, $-$ve and $0$ respectively. Similarly, the change in orientation of the robot in anti-clockwise is considered positive. 

The robot interacts with the environment uniquely for different surfaces. Consequently, the reward matrices, which indirectly model the robot-environment interactions, need to be learned for every surface and type of robot, and compensated for unexpected changes in the environment. 
\subsection{Optimization} \label{Subsection:LocomotionSequences}
A control sequence is defined as a sequence of state transitions ${S(t)}$ for  $t=0,1,\cdots,N$ . The resulting translation ($J_x,J_y$) and rotation ($J_\theta$) rewards for the given sequence are written as 
\begin{equation} \label{Eqn:J_L}
J_L \left(\{S(t)\}\right)=\sum \limits_{t=1}^N\left(T_L \right)_{S(t-1),S(t)} \quad for\ L=x,y,\theta
\end{equation}
In case of the presented robot, the four possible states can be visualized as a directed graph where elementary circuits can be computationally calculated \cite{johnson_finding_1975, hagberg_aric_networkx._????}. These elementary circuits are referred to as periodic control sequences where the first and last states are identical and no states are repeated. Exhaustively, the number of elementary circuits are $\sum \limits_{i=1}^K\left(\begin{array}{c}
K\\
K-i+1
\end{array}\right)(K-i)!$ for $K$ states \cite{johnson_finding_1975}. Let $\mathbb{E}$ represent the set of all these periodic state sequences. %
The computation of the periodic sequences and corresponding rewards facilitates  calculation of optimal control sequences for maximum translation in $+X$ direction ($S_{x}$) and rotation ($S_{\pm \theta}$)
\begin{eqnarray} \label{Eqn:S_T}
S_{x} &=& \max \limits_{\mathbb{E}}\left( J_x - (|J_y|+|J_\theta| )\right)\\ \label{Eqn:S_R}
S_{\pm \theta} &=& \max \limits_{\mathbb{E}}\left(\pm J_\theta - (|J_x|+|J_y| )\right)
\end{eqnarray}
\indent 
where $\pm$ denote the rotation in anti-clockwise and clockwise direction. 
Each control sequence results in a desired and residual (undesired) locomotion e.g. for the control sequence $S_{x},\  J_x$ is the desired translation, $J_y$ is the residual translation and $J_\theta$ is the residual rotation. As it will be shown in the next section, for the experimental soft robots with virtual grip friction mechanism, there exist two uncoupled, i.e. no residual locomotion, translation and rotation control sequences. These uncoupled sequences facilitate controlled locomotion of the robot on a planar surface.

\textit{Speed.} The variation in the speed of locomotion can either result from change in amplitude of soft body deformation or the frequency of control sequence cycles. These depend on the properties of the actuator (variation in motor power, duration of actuation), and the material properties (elasticity, rate of deformation and relaxation) of the soft material. However, the control framework evaluates the control sequences independent of these actuation and material variations. %
For example, let there be two soft robots $R_1,R_2$ with same the soft material body and identical motors at each end. The motors ${Mot}_1,{Mot}_2$ corresponding to each robot have power $P_1,P_2$ such that $P_2>P_1$. Here, the motor ${Mot}_2$, due to more power, will allow faster transition from one state to another, equivalently, higher frequency for execution of a control sequence. Hence, $R_2$ is capable of faster locomotion than $R_1$. Similar argument may also hold for robots designed using two different materials - one having faster rate of deformation than the other.

The presented control approach is flexible and adaptable. The state transition reward matrices depend on the surface of locomotion, the design of the robot and need to be learned, however, control structure remains the same. The approach is extendible to multiple limbed robots (more than two actuators and friction mechanisms) and different discretization of the control (i.e. more than two states). Additionally, the state transition rewards can be weighted alternatively e.g. normalized to robot length. Most importantly, the approach exists in task space, thus, allowing more generic applicability to robots with different actuators and materials. Despite the advantages of learning, the approach depends heavily on clever discretization of robot-environment interactions. Also, the transition reward matrix dimensions increase exponentially for multiple $M$ limbs as $P^M \times P^M$ for $P$ discretized behaviors.
\section{Experiments and Discussion} \label{Sec:Experiments}
The experimental goals are to explore the applicability of the control framework to design variations (tendon paths and friction mechanisms) and different environments. For the same, four sets of experiments are conducted involving variation of tendon path shapes, surface of locomotion, friction mechanism and motor power for observing locomotion speed. 
Soft robot bodies were 3D printed with three different tendon configurations (Fig. \ref{Fig:ManufacturedRobotConfiguration}) -– the I-shape, Midline S-shape and Simple S-shape. The two friction mechanisms - virtual grip and directional friction (Fig. \ref{Fig:DirectionalFriction}) were also printed in a modular fashion (Fig. \ref{Fig:Module}). The model-free control framework attempts to robustly capture the interaction between planar surface of locomotion and the robot friction mechanisms. The result of the interaction is stored in the set of translation and rotation transition matrices – {$T_x,T_y,T_\theta$}. The changes in the friction mechanism behavior - (\textit{discretization}) - about critical contact angle $\psi^*$ or direction of motion, are visually observed and controlled by manipulating the length of the tendons. For each robot and surface of locomotion, the \textit{learning} step is independently performed where the state transition rewards (translation and rotation) are recorded as weighted mean of the result (camera vision) of 20 repetitions of transitions between two given states (no units). Next, the set $\mathbb{E}$ containing all elementary circuits for the given robot configuration is calculated - $24$ elementary circuits for $4$ states. Finally, the \textit{optimal} periodic control sequences are selected corresponding to the locomotion task cost function using Eqs. {\ref{Eqn:J_L}}, {\ref{Eqn:S_T}}, {\ref{Eqn:S_R}}. Camera vision data is processed using Kinovea\texttrademark software for all the experiments. Experimentally, three sets of periodic control sequences were of particular interest -
\begin{eqnarray}
\mathbb{S_C} &=& \left\{S_{C1}, S_{C2}\right\}\\
&&S_{C1} = \{(10)\rightarrow (01) \rightarrow(10)\}\nonumber\\
&&S_{C2}=\{(00)\rightarrow (10) \rightarrow (01) \rightarrow(00)\} \phantom{\rightarrow (11)}\nonumber
\end{eqnarray}
\begin{eqnarray}
\mathbb{S_I} &=& \left\{S_{I1}, S_{I2}\right\} \\
&&S_{I1} = \{(10)\rightarrow (11)\rightarrow (01) \rightarrow(10)\}\nonumber\\
&&S_{I2}=\{(00)\rightarrow (10) \rightarrow (11) \rightarrow (01) \rightarrow(00)\} \nonumber
\end{eqnarray}
\begin{eqnarray}
\mathbb{S_H} &=& \left\{S_{H}\right\}\\
&&S_{H}=\{(00)\rightarrow (11) \rightarrow (00)\} \phantom{\rightarrow (00)\rightarrow (00)} \nonumber
\end{eqnarray}
For the sake of brevity during discussion, $\mathbb{S_I, S_C, S_H}$ are referred to as the inching, crawling and hopping gaits while drawing analogy from biology.

The periodic gaits exist in task-space and need to be implemented in actuator space depending upon the discretization done in Section {\ref{Subsec:Discretization}}. For the current motor-tendon actuated robots, the robot behaviors are implemented by observing the critical contact angle (Virtual Grip - Figure {\ref{Fig:VirtualGrip}} or the direction of motion (Directional Friction - Figure {\ref{Fig:DirectionalFriction}}). For the described experiments, this behavior sensing is done visually (camera), however, can be automated by augmenting each friction mechanism with an additional sensor like MEMS accelerometer.

\begin{figure}
\includegraphics[width=\columnwidth]{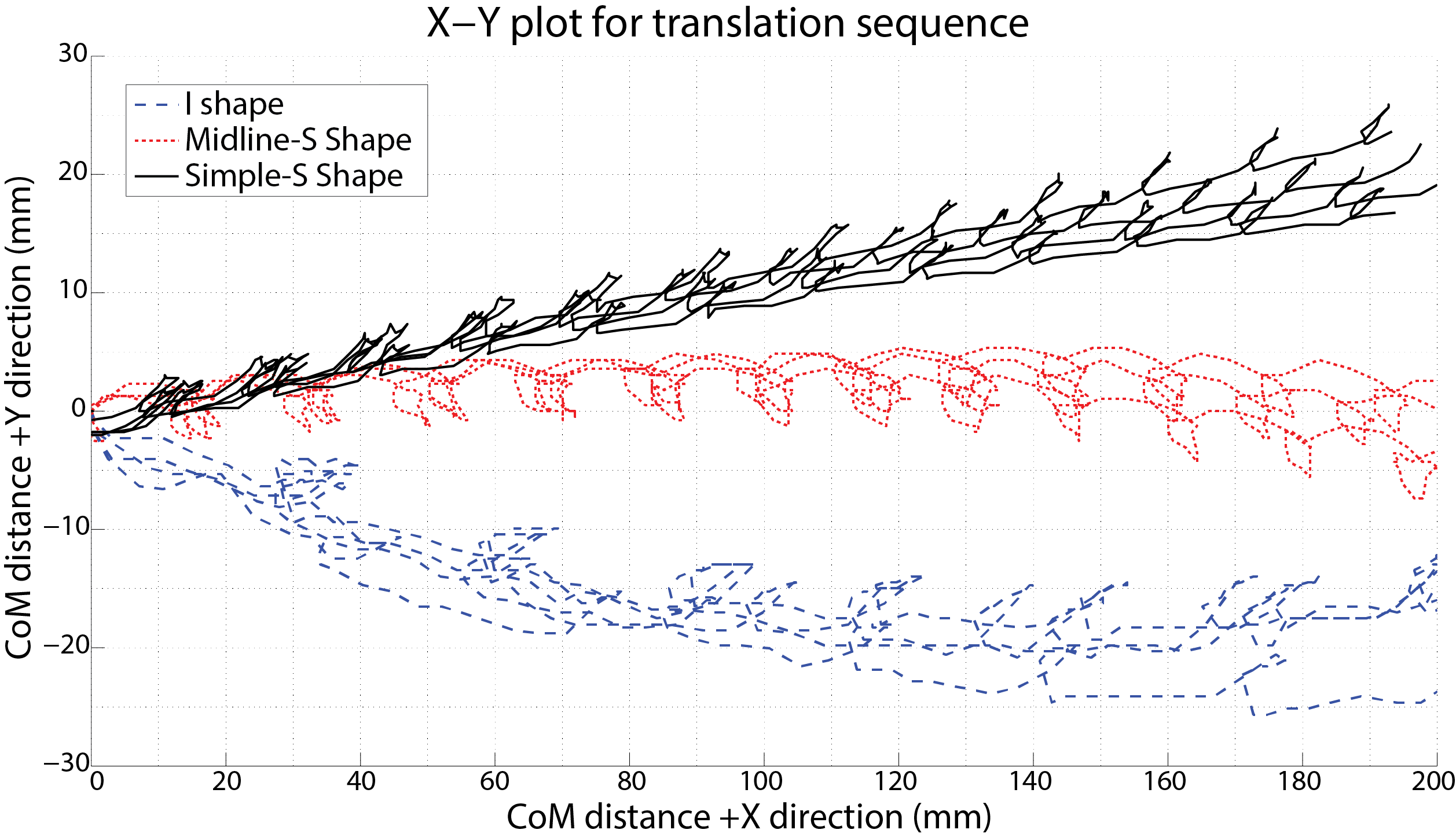}
\caption{Plot of resultant translation using the $S_{I1}$ control sequence vs the undesired residual translation ($J_y$) for the three experimental robots.}
\label{Fig:TranslationPlot}
\end{figure}

\begin{figure}
\includegraphics[width=\columnwidth]{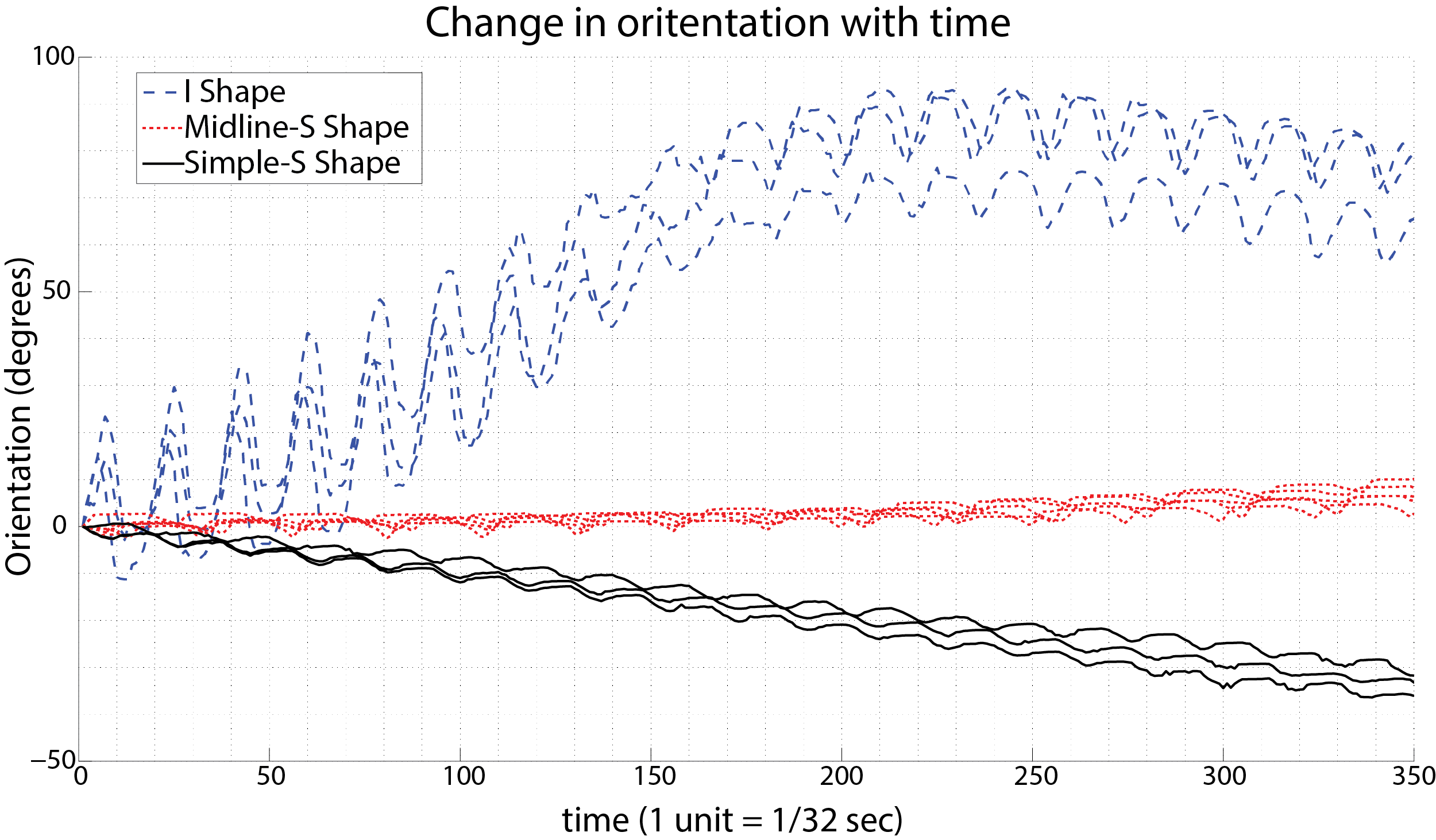}
\caption{Rotation using $S_{H}$ control sequence for the three experimental robots. The three robots display different rotational behavior – with I-shape and Simple S-shape robot displaying higher speed of rotation in anti-clockwise and clockwise direction. The Midline S-shape robot displays much slower anti-clockwise rotation.}
\label{Fig:RotationPlot}
\end{figure}

The first experiment learns the state transition reward matrices for the three tendon channel shapes as shown in Tab. \ref{Table:TransMatrices} with the virtual grip mechanism as the friction mechanism (Fig. \ref{Fig:VirtualGrip}). The optimal periodic sequence for translation was same for all the robots - $S_{I1}$ inching gait. The plot (Fig. \ref{Fig:TranslationPlot})) of the desired translation in the $+X$ direction against the residual undesired translation of the center of mass of the robot ($J_y$) illustrates the influence of tendon  configuration on the motion. Similarly, all the three tendon path shapes optimally rotate using the same hopping periodic sequence - $\mathbb{S_H}$, however, for a given motor power ($11\ V$) and control sequence cycle time ($540\ ms$), the three soft robots displayed different rotational behavior. The I-shape and Simple S-shaped robots rotated in opposite directions (Fig. \ref{Fig:RotationPlot}). The Midline S-shaped robot rotated slightly in anti-clockwise direction at a much slower speed using the $S_{H}$ control sequence. The sign of the gradient slope from the static stress analysis Fig. \ref{Fig:SlopesAndVariation} is indicative of the direction of rotation - positive for clockwise and negative for anti-clockwise direction. Hence, these results also justify the use of static stress analysis as a design tool for soft robots. 
The crawling and hopping sequences are mutually independent thereby making the soft robot capable of locomotion on a planar surface.
\begin{table}
\begin{tabular*}{\columnwidth}{@{\extracolsep{\fill}}|m{25pt}|@{}c@{}|@{}c@{}|@{}c@{}|}
\hline
 & $\mathbf{T_x}$  & $\mathbf{T_y}$ & $\mathbf{T_\theta}$ \\ \hline \hline
\includegraphics[scale=0.1, angle=90]{figures/Ishape.png} & 
{$\left[\arraycolsep=1pt%
\begin{array}{cccc}
0 & 0 & 0 & 0 \\
0 & 0 & 1 & -8 \\
0 & -1 & 0 & 8 \\
0 & 8 & -8 & 0
\end{array}\right]%
$}& 
{$\left[\arraycolsep=1pt%
\begin{array}{cccc}
0 & 0 & 0 & 0\\
0 & 0 & 0 & 1\\
0 & 0 & 0 & -1\\
0 & -1 & 1 & 0
\end{array}\right]%
$}& 
{$\left[\arraycolsep=1pt%
\begin{array}{cccc}
0 & 0 & 0 & 15 \\
0 & 0 & 0 & 0 \\
0 & 0 & 0 & 0 \\
15 & 0 & 0 & 0
\end{array}\right]%
$}\\ \hline
\includegraphics[scale=0.1, angle=90]{figures/MidlineS.png} & 
{$\left[\arraycolsep=1pt%
\begin{array}{cccc}
0 & 0 & 0 & 0 \\
0 & 0 & 2 & -8 \\
0 & -2 & 0 & 8 \\
0 & 8 & -8 & 0
\end{array}\right]%
$}& 
{$\left[\arraycolsep=1pt%
\begin{array}{c@{\hspace{7.5pt}}c@{\hspace{7.5pt}}c@{\hspace{7.5pt}}c}
0 & 0 & 0 & 0\\
0 & 0 & 0 & 0\\
0 & 0 & 0 & 0\\
0 & 0 & 0 & 0
\end{array}\right]%
$}& 
{$\left[\arraycolsep=1pt%
\begin{array}{cccc}
0 & 0 & 0 & -5 \\
0 & 0 & 0 & 0 \\
0 & 0 & 0 & 0 \\
-5 & 0 & 0 & 0
\end{array}\right]%
$}\\ \hline
\includegraphics[scale=0.1, angle=90]{figures/SimpleS.png} & 
{$\left[\arraycolsep=1pt%
\begin{array}{cccc}
0 & 0 & 0 & 0 \\
0 & 0 & 1 & -8 \\
0 & -1 & 0 & 8 \\
0 & 8 & -8 & 0
\end{array}\right]%
$}& 
{$\left[\arraycolsep=1pt%
\begin{array}{cccc}
0 & 0 & 0 & 0\\
0 & 0 & 1 & -1\\
0 & -1 & 0 & 1\\
0 & 1 & -1 & 0
\end{array}\right]%
$}& 
{$\left[\arraycolsep=1pt%
\begin{array}{cccc}
0 & 0 & 0 & -15 \\
0 & 0 & 0 & 0 \\
0 & 0 & 0 & 0 \\
-15 & 0 & 0 & 0
\end{array}\right]%
$} \\ \hline
\end{tabular*}\\
\caption{State transition reward matrices for the three different tendon channel designs for the soft robot moving on a smooth planar surface. They capture the interaction between the surface and discretized friction mechanism action.}
\label{Table:TransMatrices}
\end{table}

The second experiment learns the state transition matrices on two different surfaces and analyzes the behaviors. The robot was Midline S-Shape tendon shape with virtual grip friction mechanism and $11V$ actuation. The Table {\ref{Table:TransMatricesSurface}} compares state transition reward matrices for robot behavior on a smooth texture table and a rough texture office carpet. The interesting observation was that the robot displayed optimal translation using an inch $\mathbb{S_I}$ gait on table top while preferring crawl $\mathbb{S_C}$ gait on a rougher carpet surface. However, the optimal rotation was achieved using $\mathbb{S_H}$ hop gait in both cases.
\begin{table}
\begin{tabular*}{\columnwidth}{@{\extracolsep{\fill}}|c|c|c|}
\hline
 & \textbf{Fine textured table}%
 & \textbf{Rough textured carpet} \\ \hline \hline
$\mathbf{T_x}$ & 
{$\left[\arraycolsep=2pt%
\begin{array}{cccc}
0 & 0 & 0 & 0 \\
0 & 0 & 2 & -8 \\
0 & -2 & 0 & 8 \\
0 & 8 & -8 & 0
\end{array}\right]%
$}& 
{$\left[\arraycolsep=2pt%
\begin{array}{cccc}
0 & 0 & 0 & 0 \\
0 & 0 & -7 & 1 \\
0 & 7 & 0 & 5 \\
0 & -1 & -5 & 0
\end{array}\right]%
$}\\ \hline
$\mathbf{T_y}$ & 
{$\left[\arraycolsep=2pt%
\begin{array}{cccc}
0 & 0 & 0 & 0 \\
0 & 0 & 0 & 1 \\
0 & 0 & 0 & -1 \\
0 & -1 & 1 & 0
\end{array}\right]%
$} & 
{$\left[\arraycolsep=3pt%
\begin{array}{cccc}
0 & 0 & 0 & 0 \\
0 & 0 & 0 & 1 \\
0 & 0 & 0 & -1 \\
0 & -1 & 1 & 0
\end{array}\right]%
$}\\ \hline
$\mathbf{T_\theta}$ & 
{$\left[\arraycolsep=2pt%
\begin{array}{cccc}
0 & 0 & 0 & -5 \\
0 & 0 & 0 & 0 \\
0 & 0 & 0 & 0\\
-5 & 0 & 0 & 0
\end{array}\right]%
$}& 
{$-\left[\arraycolsep=1pt%
\begin{array}{cccc}
0 & 0 & 0 & 30 \\
0 & 0 & 5 & 15 \\
0 & 5 & 0 & 15 \\
30 & 15 & 15 & 0
\end{array}\right]%
$} \\ \hline
\end{tabular*}\\
\caption{State transition reward matrices for robots with different locomotion surfaces - fine textured table (column 2) and rough textured carpet (column 3).}
\label{Table:TransMatricesSurface}
\end{table}
%

The third experiment observed the application of the control framework to a different friction mechanism - direction friction (Fig. \ref{Fig:DirectionalFriction}). Here, two cases are considered - symmetrically and non-symmetrically placed directional friction mechanisms for a I-shape soft robot body. For the symmetrically placed case, tension of the tendon (coiling) is the preferred direction for the friction mechanisms at both ends of the robot. While, for the non-symmetrical case - tension of tendon is the preferred direction for the rear, while relaxation of the tendon is the preferred direction for the front mechanism as visible in the first row of Tab. {\ref{Table:TransMatricesDF}}. Here, the hollow arrows indicate the preferred direction of motion of the friction mechanism. The table presents the learned transition matrices for the two robots.  The optimal periodic translation sequences were identified as $S_{I1}$ inch gait and $S_{C1}$ for symmetric and non-symmetric cases. However, the rotational residual motion ($J_\theta$) is more substantial for the former case. The intuitive unidirectional translation in the non-symmetrical case is captured in the translation matrix $T_x$. The deformation motion of the soft body resulting in translation (forward moving wave) is similar for the virtual grip and directional friction robots but the control sequences (equivalent motor activation) are different and can be obtained by applying the same framework to all these robots. 

\begin{table}
\begin{tabular*}{\columnwidth}{@{\extracolsep{\fill}}|c|c|c|}
\hline
 &\includegraphics[scale=0.3]{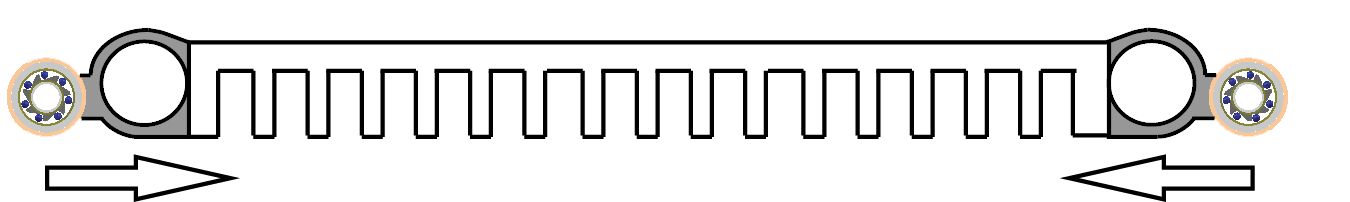} %
 &\includegraphics[scale=0.3]{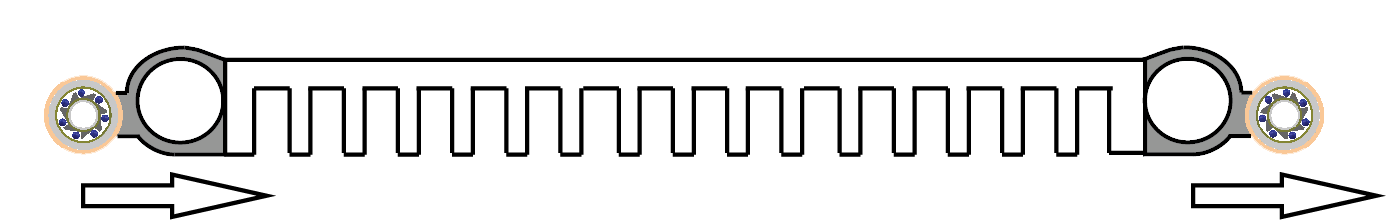} \\ \hline \hline
$\mathbf{T_x}$ & 
{$\left[\arraycolsep=2pt%
\begin{array}{cccc}
0 & -2 & 2 & 0 \\
2 & 0 & 2 & 0 \\
-2 & -2 & 0 & 0 \\
0 & 0 & 0 & 0
\end{array}\right]%
$}& 
{$\left[\arraycolsep=2pt%
\begin{array}{cccc}
0 & 0 & 2 & 2 \\
0 & 0 & 5 & -2 \\
2 & 5 & 0 & 0 \\
2 & 2 & 2 & 0
\end{array}\right]%
$}\\ \hline
$\mathbf{T_y}$ & 
{$\left[\arraycolsep=1pt%
\begin{array}{cccc}
0 & 0 & 0 & 0 \\
0 & 0 & 0.5 & 0 \\
0 & -0.5 & 0 & 0 \\
0 & 0 & 0 & 0
\end{array}\right]%
$}& 
{$\left[\arraycolsep=3pt%
\begin{array}{cccc}
0 & 0 & 0 & 0 \\
0 & 0 & 0 & 0 \\
0 & 0 & 0 & 0 \\
0 & 0 & 0 & 0
\end{array}\right]%
$}\\ \hline
$\mathbf{T_\theta}$ & 
{$-\left[\arraycolsep=2pt%
\begin{array}{cccc}
0 & 10 & 10 & 10 \\
10 & 0 & 10 & 5 \\
10 & 10 & 0 & 5 \\
10 & 5 & 5 & 0
\end{array}\right]%
$}& 
{$\left[\arraycolsep=1pt%
\begin{array}{cccc}
0 & -1 & 1 & -1 \\
-1 & 0 & 0 & 0 \\
1 & 0 & 0 & 0 \\
-1 & 1 & 0 & 0
\end{array}\right]%
$} \\ \hline
\end{tabular*}\\
\caption{State transition reward matrices for robots with directional friction mechanisms but oriented in symmetric (column 2) and non-symmetric (column 3) manner. The hollow arrows indicate the preferred direction of translation (low friction).}
\label{Table:TransMatricesDF}
\end{table}

\begin{figure}
{\centering \includegraphics[width=\columnwidth]{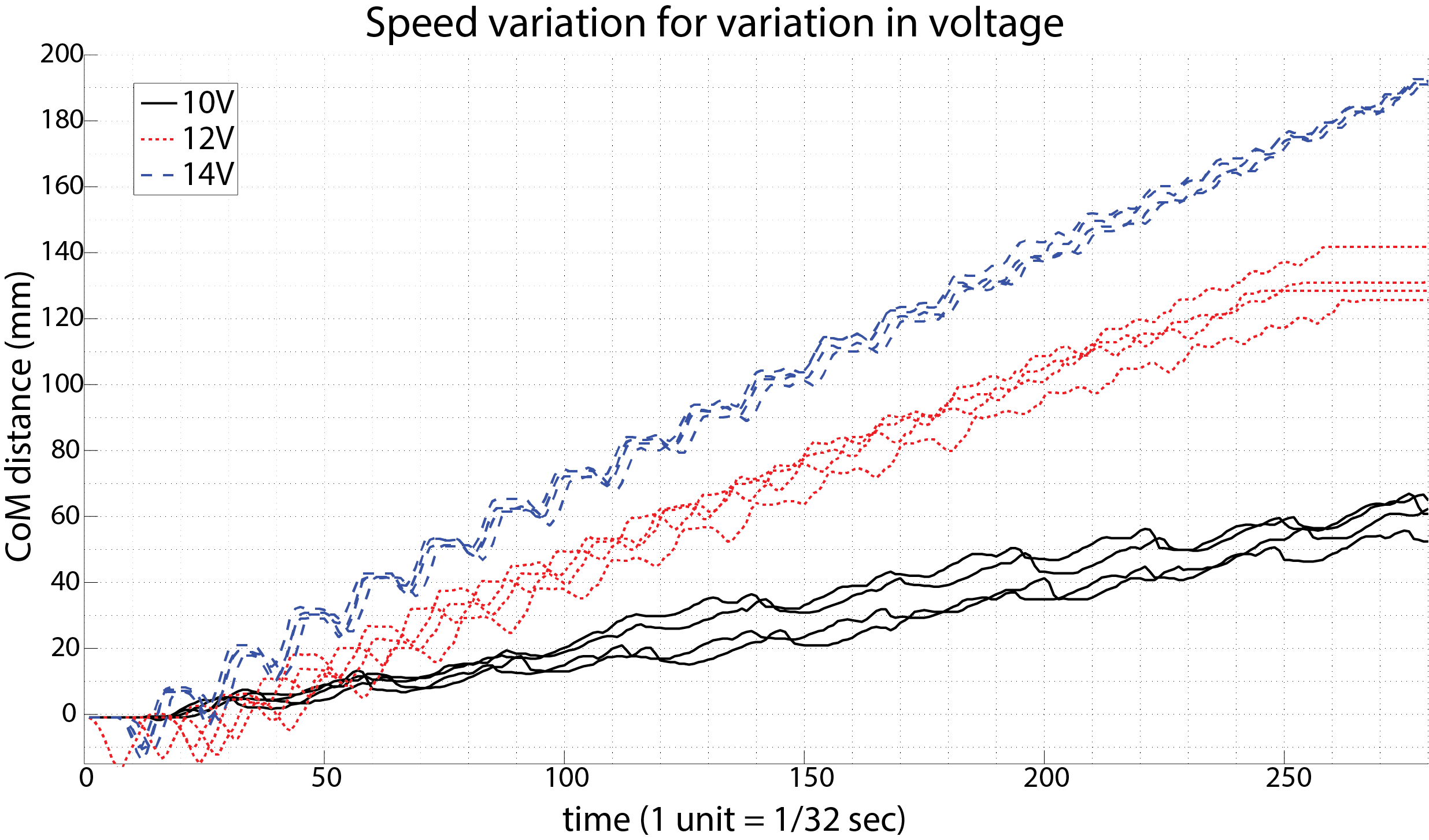}
\subcaption{Symmetrically placed virtual grip mechanisms for Midline S-shape robot body.}
\label{Fig:Speed_VG_MSshape}}
{ \centering
\includegraphics[width=\columnwidth]{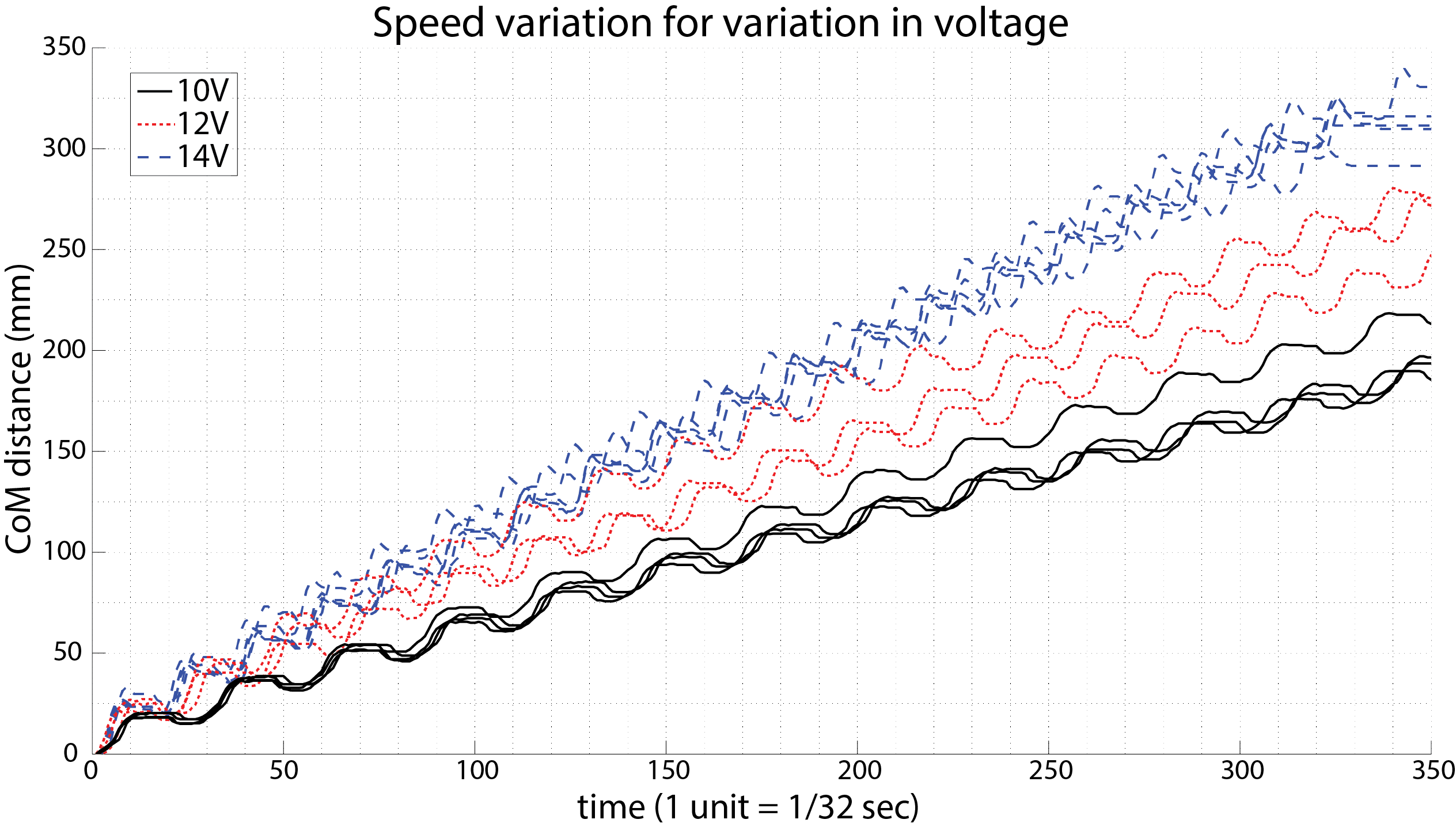}
\subcaption{Symmetrically placed directional friction mechanisms for I-shape robot body.} \label{Fig:Speed_DF_Ishape}}
\caption{Comparison of translation of center of mass with varying motor power. The magnitude of deformation per cycle and the frequency of control sequence cycles increases with the voltage.} \label{Fig:Speed}
\end{figure}

The final experiment analyzes the effect of power on translation speed of the robot with the intention to explore time-independence of the current control framework i.e. the capabilities of actuator-material combination to move from one robot state to another. Two robots utilized in this experiment are - 1) Midline-S shape body with symmetrically placed virtual grip friction mechanisms (Fig. \ref{Fig:Speed_VG_MSshape}), and 2) I-shape body with symmetrically placed directional friction mechanisms (Fig. \ref{Fig:Speed_DF_Ishape}). The robots were powered at 3 different voltages - $10V,\ 12V,\ 14 V$ to execute the $S_{I1}$ inching periodic sequences while the center of mass was visually tracked. For the given soft material, higher voltage actuation enables faster transition from one state to another, thus, higher frequency of periodic control cycles. Additionally, the deformation of the body per cycle is also higher. This is analogous to increase in speed of walking due to faster steps (higher cycle frequency) and larger step size (more body deformation per cycle).

The experiments provide insight into the application of the learning-based model-free control framework and soft robot design. Static stress analysis was successfully used as a tool to design motor-tendon actuators for soft body robots. From design perspective, 3D printing proved to be a very useful prototyping tool as it facilitates exploration of complex designs with multi-materials, however, limits the variety materials that can be utilized. The model-free control scheme proved to be effective where learning is an important step and facilitates calculation of periodic control sequences, however, currently it does not use any prior knowledge of robot morphology.
\section{Conclusion and Future Work}
The research introduces a 3D-printed soft robot driven with a motor-tendon combination capable of terrestrial locomotion. The rapid prototyped modular robot design (separate soft body and friction mechanisms) facilitates rapid fabrication, deployment and repair. The manipulation of friction to facilitate locomotion is performed using two friction manipulation mechanisms - the virtual grip mechanism and the directional friction mechanism. Additionally, the motor-tendon actuators are consistent in activation. 

The design methodology for motor-tendon driven soft robots discusses the use of static stress analysis as a design tool for tendon path shapes. The modulus disparity between the tendon and soft body is overcome by a design conceptually similar to that of bowden cables by printing hard material shell along the tendon path. The deformation of the soft body results in stress concentration at the end of the tendon path where it attaches to the body and directly affects the soft body-hard shell interface. This stress concentration can be distributed by designing gradient boundaries between the materials.

The concept of model-free control for soft robots is presented and experimentally applied to the robot designs with different tendon path shapes, friction mechanisms and variable environments. The model-free control framework discretizes the factors that dominate the robot-environment interaction (friction mechanism behaviors) and in process defines finite robot states. The resulting transitions from one state to another are learned and stored in state transition matrices. Next, the periodic control sequences are calculated using elementary circuits in directed graphs. Finally, optimized periodic control state sequence is calculated corresponding to a desired task cost function (translation or rotation). These sequences exist in task-space and are implemented on specific robot corresponding to the state discretization i.e. direction of motion for directional friction mechanism and contact angle variation for virtual grip. For a given robot, the speed of the robot is determined by the rate at which the robot can transition from one state to another. Conceptually, it is independent of actuator, material and even type of friction mechanism. In the first experiment, the three experimented tendon paths designs display decoupled translation and rotation behaviors using crawling $\mathbb{S_C}$ and hopping $\mathbb{S_H}$ gaits. The change in shape of the tendon paths is reflected in the rotation (direction and speed) of the different robots – clockwise for I-shape and anti-clockwise for Simple S-shape tendon channel robots. This can be viewed as motivation to program computation into the robot morphology.

From design perspective, the adaptation of motor-tendon cable and different friction mechanisms to soft robots is foreseen to be instrumental in design of more sophisticated semi-autonomous soft robots for terrestrial locomotion. The use of 3D printing as a prototyping tool facilitates more complex designs and faster manufacturing. The formalization of control framework for multi-limb robots will help in more generic application to discrete point contact robots. The use of probabilistic state transition rewards as opposed to current deterministic rewards is expected to be explored in future to facilitate better understanding of the environment. The research provides evidence for evolution of the control framework and soft robot designs that will allow locomotion of soft robots in unstructured environments.
\section{Acknowledgement}
The authors would like to thank Dr. Takuya Umedachi for sharing resources for the experiments. This work was funded in part by the National Science Foundation grant IOS-1050908 to Barry Trimmer and National Science Foundation Award DBI-1126382.
%
\bibliographystyle{ieeetr}
\bibliography{softrefs}
%
\end{document}